\documentclass[twocolumn]{aastex62}

\usepackage{amsmath}
\usepackage{amsthm}
\usepackage{amsfonts}
\usepackage{amssymb}
\usepackage{enumerate}
\usepackage{asymptote}
\usepackage{float}
\usepackage{wrapfig}

\usepackage{graphics}
\usepackage{graphicx}
\usepackage{epsfig}
\usepackage{color}
\usepackage{verbatim}

\shorttitle{Leveraging Sequence Models to Predict the Useful Life of Batteries}
\shortauthors{Paradis \& Whitmeyer}

\begin{document}


\title{\Large \textbf{Pay Attention: Leveraging Sequence Models to Predict the Useful Life of Batteries}}


\correspondingauthor{Samuel Paradis}
\email{samparadis@berkeley.edu}

\correspondingauthor{Michael Whitmeyer}
\email{mwhitmeyer@berkeley.edu}

\author{Samuel Paradis}
\affil{Department of Electrical Engineering and Computer Sciences, University of California, Berkeley, CA 94720}

\author{Michael Whitmeyer}
\affil{Department of Electrical Engineering and Computer Sciences, University of California, Berkeley, CA 94720}







\begin{abstract}
We use data on 124 batteries released by Stanford University to first try to solve the binary classification problem of determining if a battery is ``good" or ``bad" given only the first 5 cycles of data (i.e., will it last longer than a certain threshold of cycles), as well as the prediction problem of determining the exact number of cycles a battery will last given the first 100 cycles of data. We approach the problem from a purely data-driven standpoint, hoping to use deep learning to learn the patterns in the sequences of data that the Stanford team engineered by hand. For both problems, we used a similar deep network design, that included an optional 1-D convolution, LSTMs, an optional Attention layer, followed by fully connected layers to produce our output. For the classification task, we were able to achieve very competitive results, with validation accuracies above 90\%, and a test accuracy of 95\%, compared to the 97.5\% test accuracy of the current leading model. For the prediction task, we were also able to achieve competitive results, with a test MAPE error of 12.5\% as compared with a 9.1\% error achieved by the current leading model \citep{Severson}.
\end{abstract}


\section{Introduction}

In this paper we aim to tackle the problem of predicting how long a lithium-ion battery will last before it ``dies", meaning it reaches approximately 80\% of its original capacity. There are many factors which go into a battery's state of health (SOH), and a lot of complex chemical processes \citep{Brissota, Vettera}. Dendrites can build up in the battery, and a phenomenon known as SEI layer growth can both lead to degradation in battery capacity, but both have been difficult to accurately and quantitatively model with known equations \citep{Vettera}. This has led some to try a semi-empirical or fully empirical approach to battery aging. Recently, Stanford and MIT released a joint effort attempting to predict how many \textit{cycles} a battery would last, where a cycle is simply a full charge and discharge of the battery \citep{Severson}. They tried to predict: 

1. Given data on the first 5 cycles, predict whether the battery will last past a certain cycle threshold, which is a binary classification problem.

2. Given data on the first 100 cycles, predict exactly how many cycles the battery will last. 

They attempted to solve these problems by engineering features and performing regularized linear regression in the engineered feature space. We will attempt to solve the problem by using deep learning architectures that can process sequences of data in order to allow relevant patterns and features \textit{to be learned} rather than hand-selected. 

Solving the problem of battery degradation uncertainty is important because a solution has the potential to get new battery technologies to the market faster, save materials/energy, and allow consumers to have access to more consistently reliable batteries. To get an accurate understanding of the cycle-life of a battery, currently you must drain thousands of these batteries to failure. Further, every batch of batteries has outliers: a small percentage of batteries produced will have significantly shorter cycle-life than expected. Is there a way we could classify and predict the cycle-life of a Lithium-ion battery with deep learning to both analyze battery life and identify outliers, without actually ever draining the batteries themselves, which wastes materials and energy? 

In order to answer this question, we will split the dataset of 124 batteries into a training set, validation set, and test set. For classification, the accuracy will simply be the percentage of correct classifications. For predicting the exact cycle-life, the accuracy will be in terms of the mean absolute percentage error (MAPE), which is the metric often used to measure success in this context. We hoped to achieve accuracies comparatively competitive to \cite{Severson}, which represents the current leading model for battery degradation. 

\section{Data}

The data we used is associated with the paper 'Data driven prediciton of battery cycle life before capacity degradation' by \cite{Severson}. The researchers have made the dataset--the largest of its kind--publicly available. The data consists of 124 batteries, with each battery lasting a certain number of cycles ($\sim$1000), and each cycle containing data on various measurements, such as temperature, voltage, current, and charge/discharge capacity for each time step of the cycle. Processing the data involved extracting information from a set of nested dictionaries, and then interpolating or zero padding where necessary to ensure the data matrix would be a tensor. Cleaning the data involved developing a simple outlier removal tool to remove noisy readings. An outline of the variables observed over the course of each cycle is in Table 1:

\begin{deluxetable}{cl}[H]
\tablecaption{Relevant Variables in Dataset}
\tablecolumns{2}
\tablehead{
\colhead{Variable} & 
\colhead{Description} 
}
\startdata
\textit{T} & Temperature  \\
\textit{V} & Voltage  \\
\textit{Qd} & Discharge Capacity  \\
\textit{Qd-lin} & Linearly interpolated discharge capacity (over cycle)  \\
\textit{Td-lin} & Linearly interpolated temperature (over cycle)  \\
\textit{dQ/dV} & Change in discharge capacity over change in voltage  \\
\enddata
\tablecomments{We define an attribute as the mean, variance, minimum, or maximum of a variable over an entire cycle.}

\end{deluxetable}

A key problem we faced is that we  wanted our model to learn complex sequential patterns, but the dataset is quite small, with only 124 $(X,Y)$ data-label pairs. Each of the batteries had hundreds of cycles, and each cycle had thousands of data points, but we found that much of the data was 1) very similar looking 2) not useful for prediction. With the model initially failing to learn when passing in all of the possible data within each cycle, we determined that we needed to reduce the data (via mean, variance, max, min), and then identify the attributes that provide meaningful insights for the model. In order to do so, we visualized each variable. These plots are relevant because at each timestep, a battery contains a significant amount of information, and in order to optimize the ability of the LSTM-based models to learn, we needed to pass in the most relevant data. We acknowledge that filtering noise is a strength of deep networks; in this specific instance, due to the sparcity of the data, feature engineering was required to attain any meaningful results. 

Each point on the plot represents the cycle-life of a battery as function of the change between cycle 100 and cycle 10 of some reduction function over all the data for some variable collected during the cycle. These plots identify the variables that provide meaningful information. Looking at the plots, any plot where the data is spread uniformly vertically conveys little information for predicting cycle-life; the cycle life is independent from the attribute in this case. 

\begin{widetext}

\begin{figure}[H]
\title{\textbf{Very Useful Attributes}}
\centering

\includegraphics[width=220px]{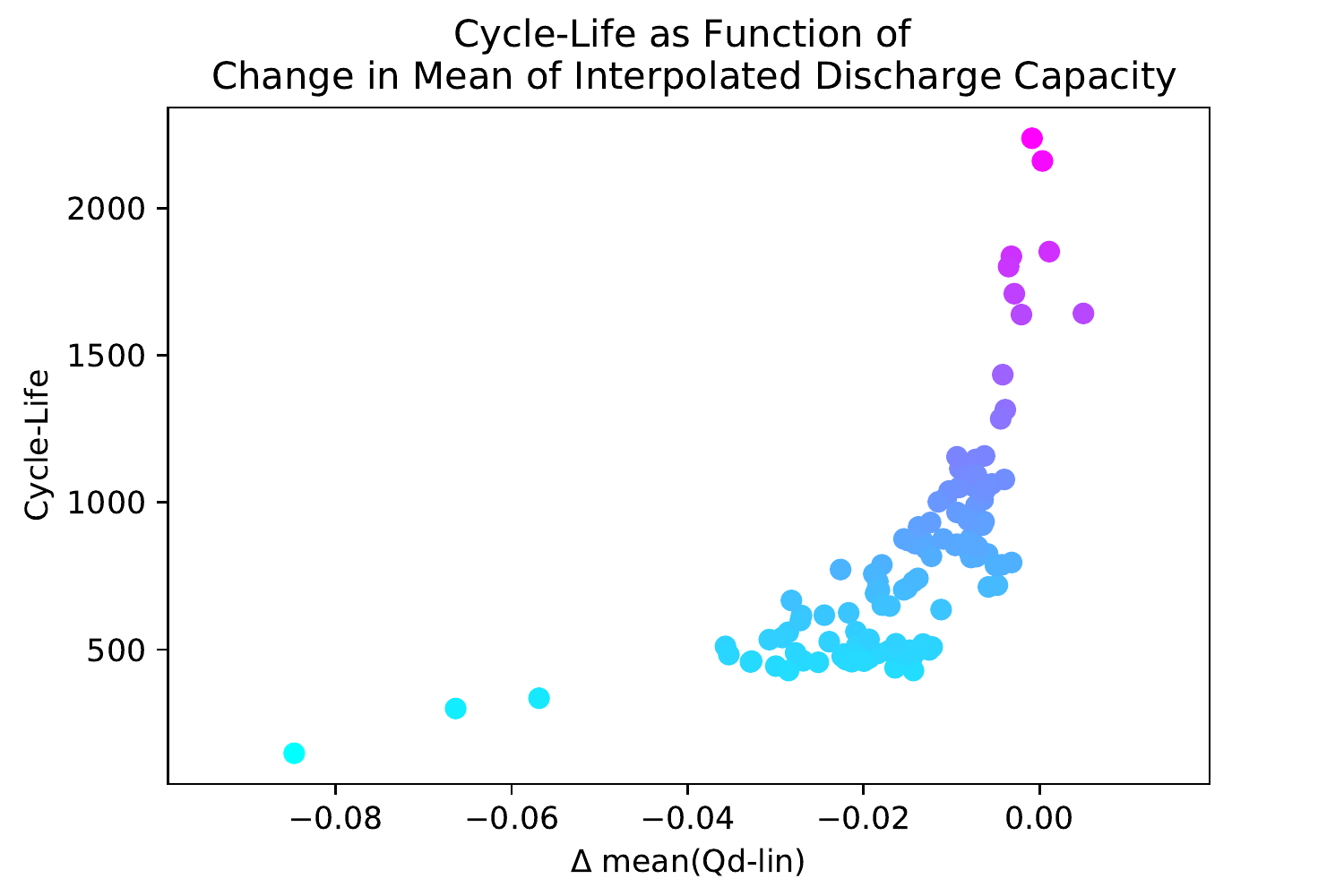}
\includegraphics[width=220px]{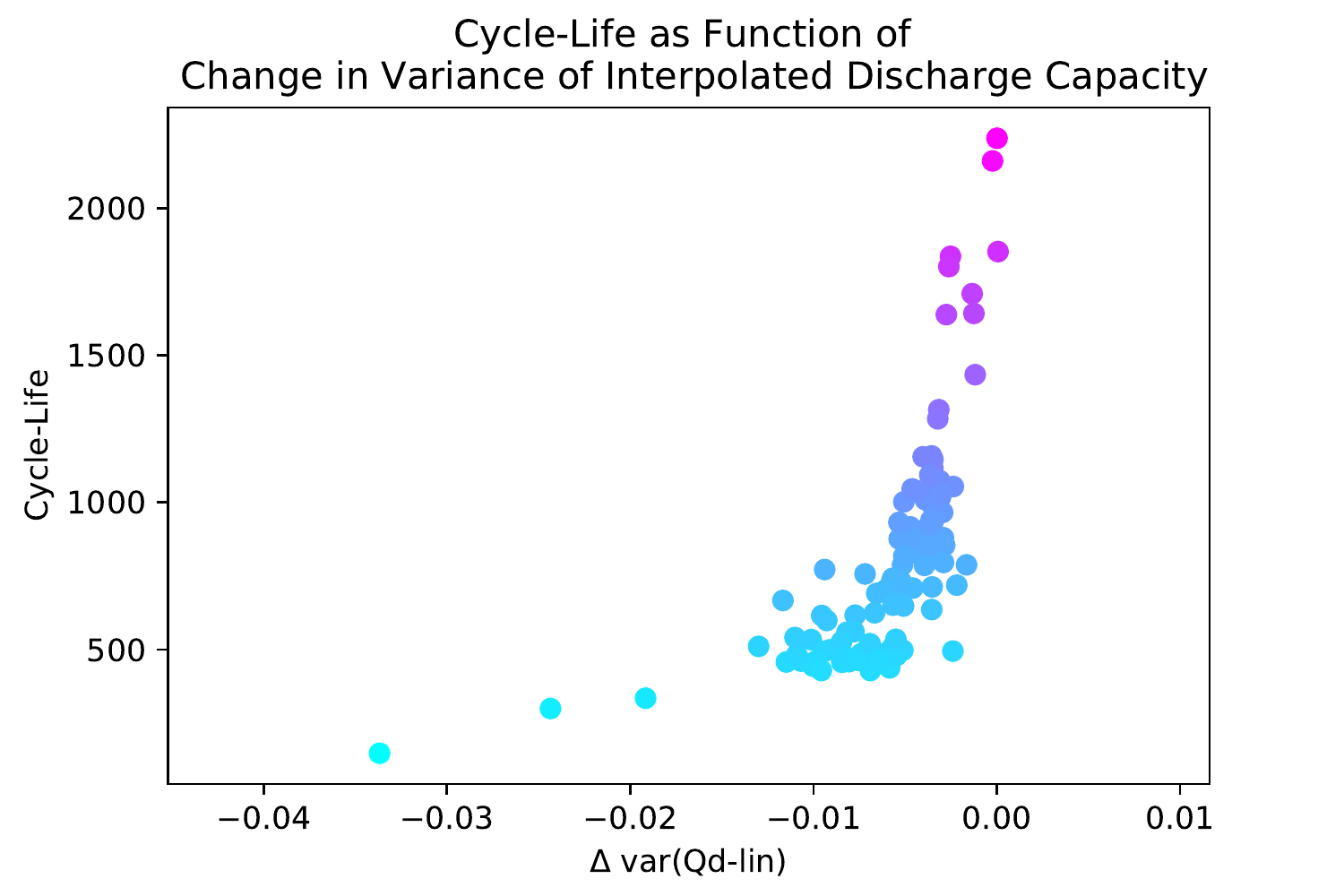}
\caption{Top ranked attributes: mean(\textit{Qd-lin}), var(\textit{Qd-lin}), var(\textit{dQ/dV}). For these attributes, we see the value varied between cycles, and this change directly impacted the cycle-life.}
\end{figure}
\begin{figure}[H]
\title{\textbf{Somewhat Useful Attributes}}
\centering

\includegraphics[width=220px]{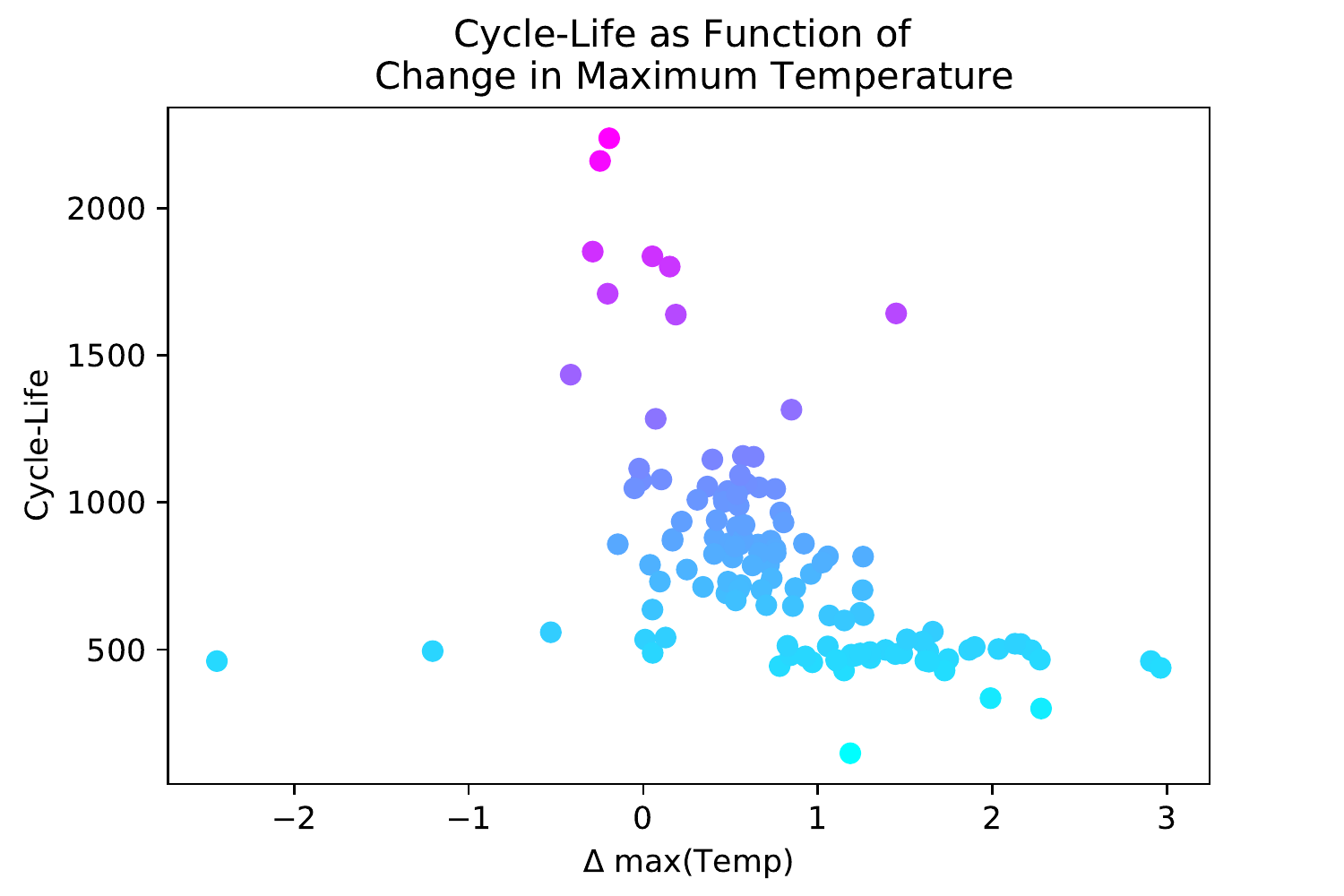}
\includegraphics[width=220px]{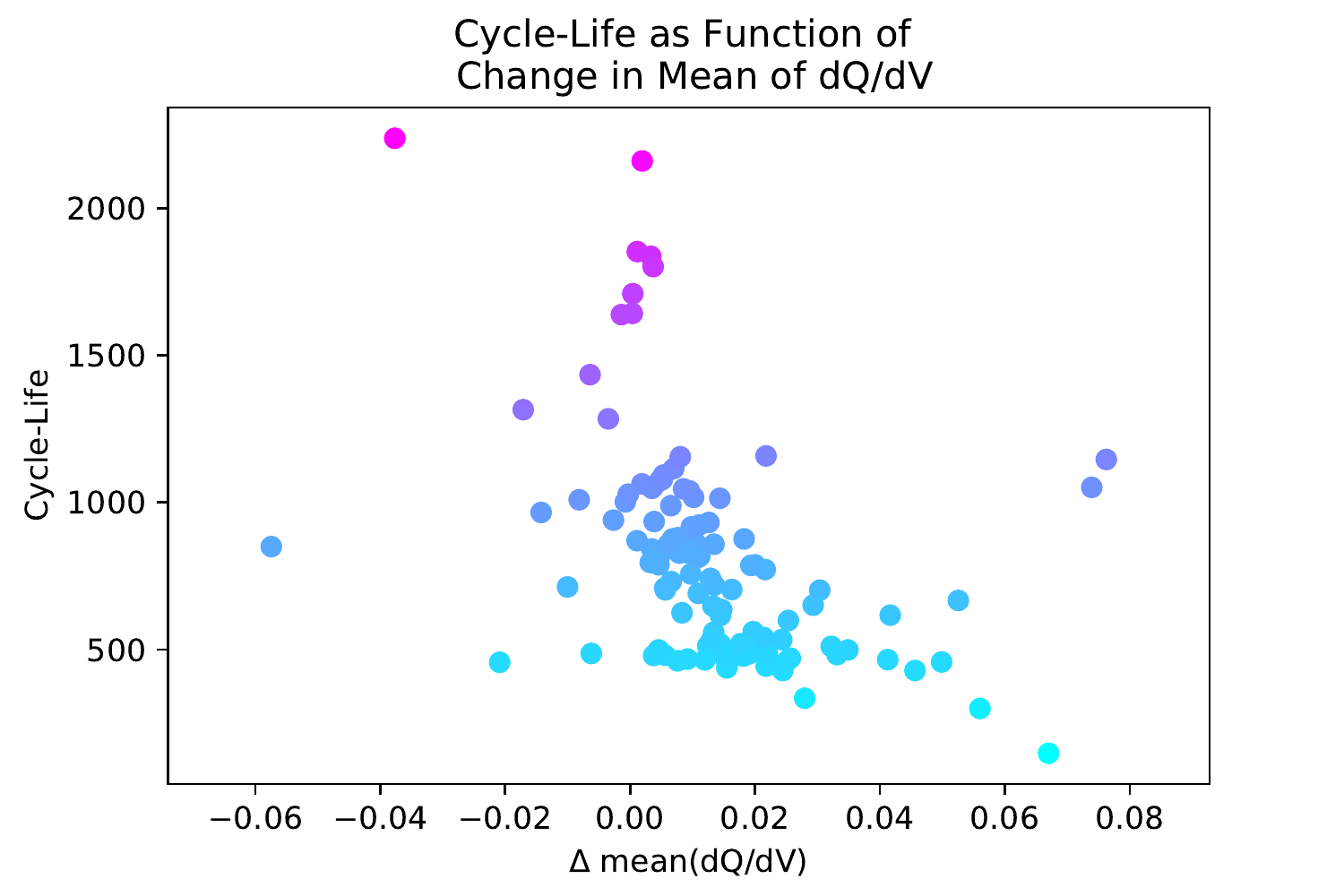}
\caption{Middle ranked attributes: max(\textit{T}), mean(\textit{dQ/dV}), mean(\textit{Qd}), max(\textit{Qd-lin}), min(\textit{T}), var(\textit{T}), mean(\textit{Td-lin}), max(\textit{Qd}), mean(\textit{T}), max(\textit{Td-lin}), var(\textit{Td-lin}), min(\textit{Td-lin}). For these attributes, we see the value varied between cycles, and this variation impacted the cycle-life.}
\end{figure}
\begin{figure}[H]
\title{\textbf{Not Useful Attributes}}
\centering

\includegraphics[width=220px]{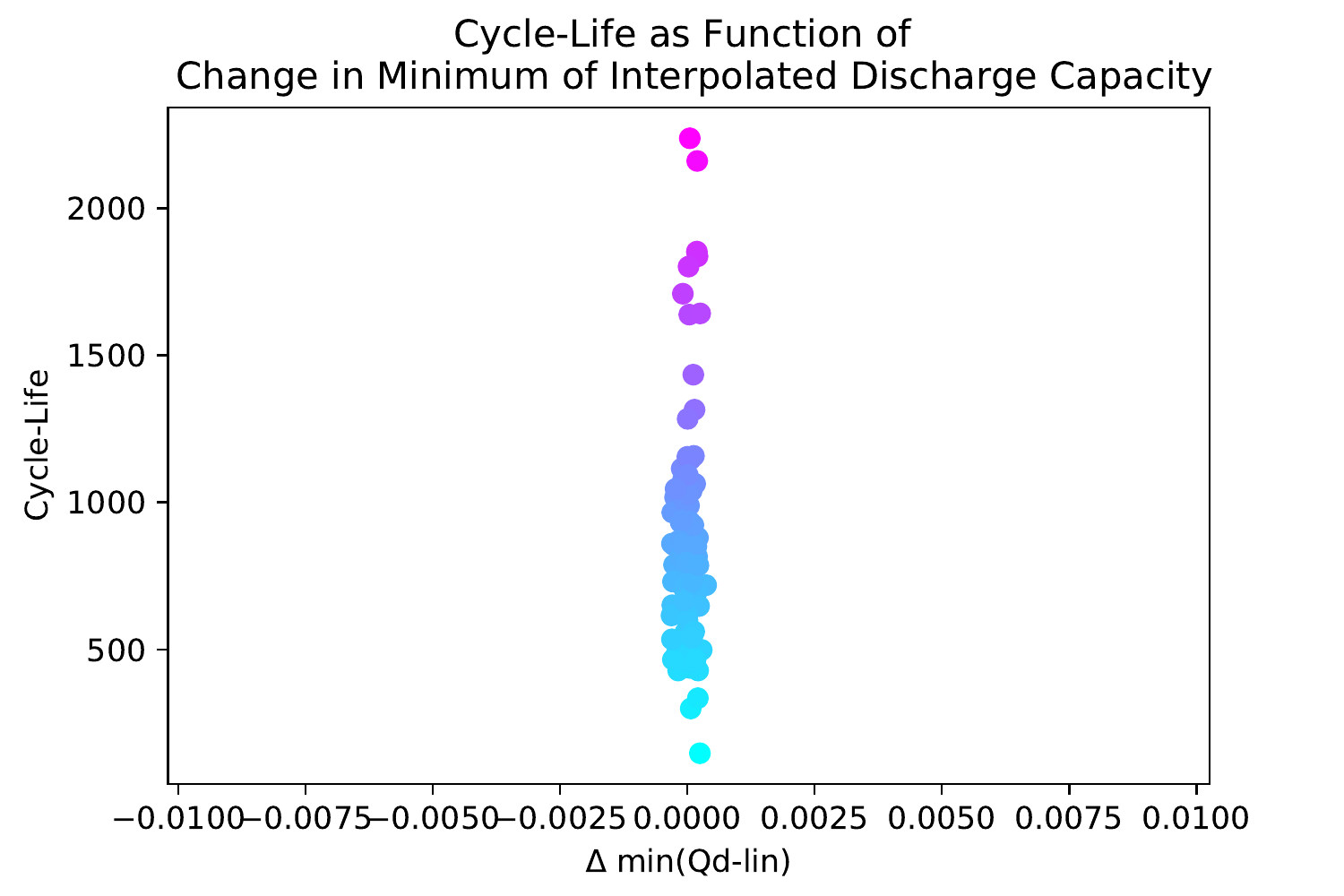}
\includegraphics[width=220px]{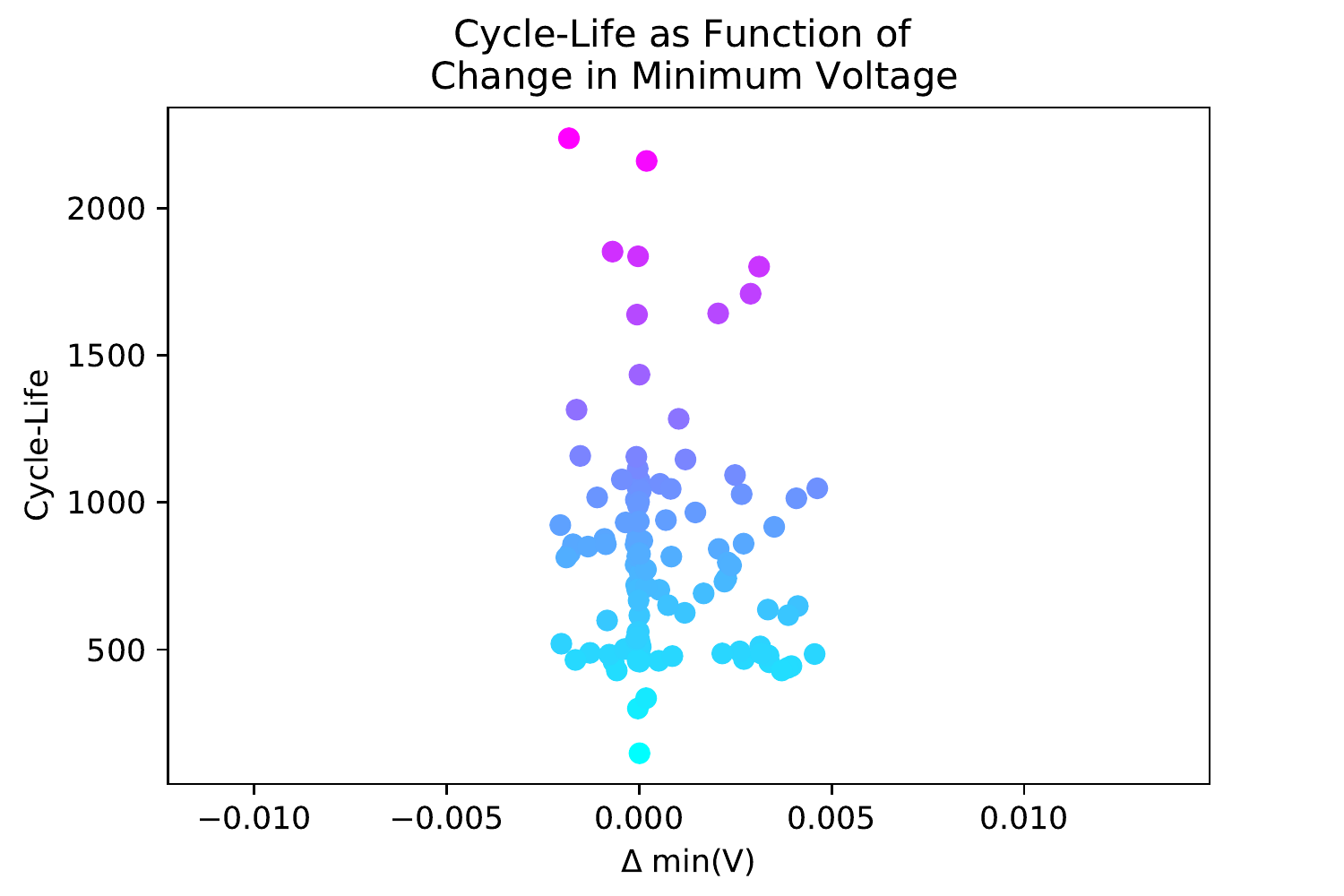}
\caption{Lowest ranked attributes: min(\textit{Qd-lin}), min(\textit{V}), min(\textit{Qd}), max(\textit{V}), mean(\textit{V}), var(\textit{V}), min(\textit{dQ/dV}), max(\textit{dQ/dV}), var(\textit{Qd}). For these attributes, we see the value fails to vary between cycles, and when it does, this variation does not seem to impact the cycle-life.}
\end{figure}

The input to our model was the top and middle ranked attributes of the first 5 cycles for classification and first 100 cycles for prediction, passed in sequentially by cycle.

\end{widetext}

\section{Models}

While previous attempts relied on the physical properties of batteries, as well as human trial and error, to identify the relevant sequential relations between cycles, our sequence model allowed for these sequential relations to be learned. The current leading model processes the data by passing in the differences between certain values in the 5$^{\text{th}}$ and 1$^{\text{st}}$ cycle for classification, and 100$^{\text{th}}$ and 10$^{\text{th}}$ cycle for prediction \citep{Severson}. Instead of using differences, we wanted to leverage LSTMs and Attention to process the data of the cycles sequentially to allow relevant patterns be learned rather than hand-selected. The output of this sequence section is passed into dense layers to both increase the flexibility of the model to fit the data, as well as to reduce the output to a single number. 

\begin{figure}[H]
\centering
\includegraphics[width=.82\linewidth]{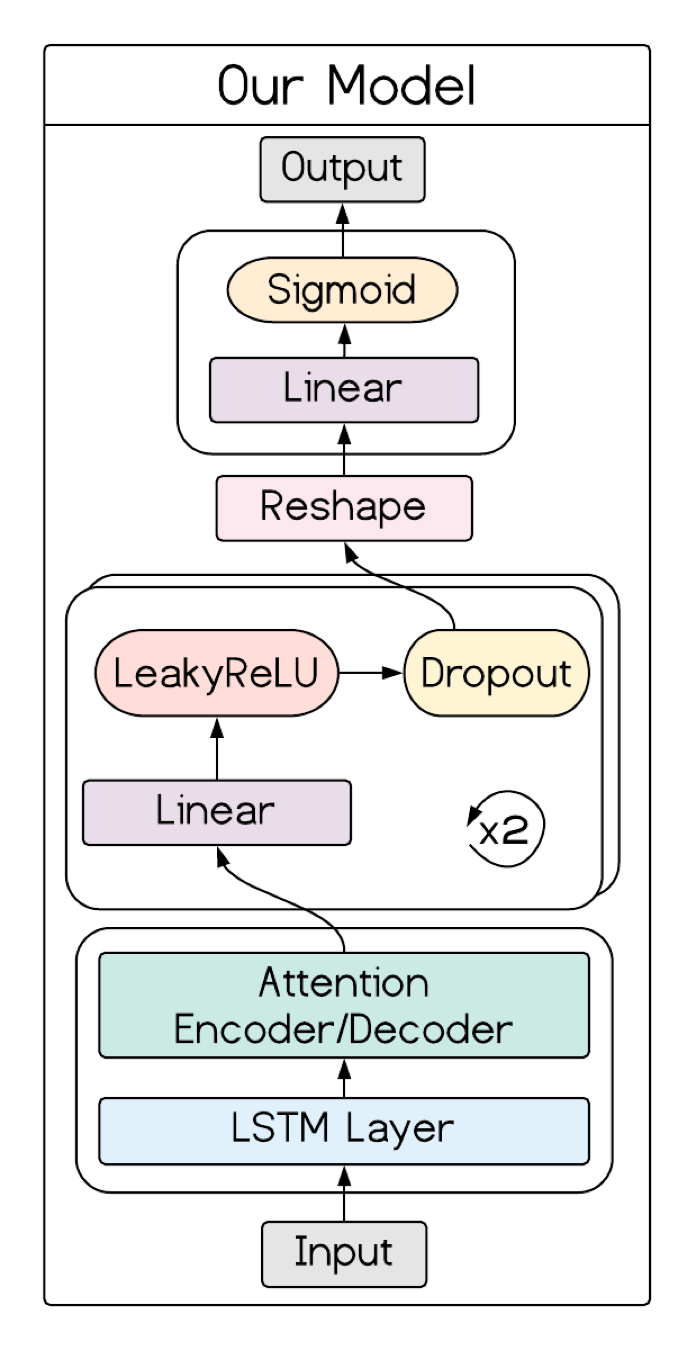}
\caption{Baseline Model. Note the Attention Encoder/Decoder is optional, and the sigmoid activation was only used for the classification task.}
\end{figure}

\subsection{Classification}

For the classification problem, we tuned the baseline model to maximize performance for the classification task. This involved iterating over different models and evaluating them based on their training and validation performance. We investigated models with LSTM/Attention hidden vector sizes ranging from 16 to 256, dense layers with sizes ranging from 16 to 256, with and without dropout, and also with and without a second and third hidden layer. We found that the more complex models struggled with overfitting the training data, which makes intuitive sense considering the small size of the dataset. Validation accuracy was low for many of the complex models, especially when the LSTM size was at or above 128. In order to allow the model to generalize well, we used a simple form of our base model: 
\begin{itemize}
\item LSTM/Attention hidden size of 16, followed by two dense layers with 16 outputs and one dense layer with one output.
\item Used dropout with .3 probability and LeakyReLU between the dense layers, ending with a sigmoid activation. 
\item Table 2 shows the training, validation, and test classification accuracy for this model, with Figure 6 visualizing the results on the test set.


\end{itemize}

\subsection{Prediction}

In order to predict the exact number of cycles a battery would last, we followed \cite{Severson}'s example and used the first 100 cycles of data as our input. We experimented with using summary data (for example, the mean of the temperature over the cycle), as opposed to inputting all the data available within a cycle, but we once again found that summarizing the data in a smart way was more effective than trying to throw all of our data (the vast majority of which was not useful for prediction) at our model, and hoping it learned what was useful and what was not. However, we also found that it was very useful to add a 1-D convolutional layer before feeding our sequence of 100 cycles into our LSTM. The reasoning behind this is that LSTMs (despite their name), can be quite forgetful, and learning a sequence of 100 inputs can be difficult to train, especially with our limited amount of data. We also considered using some sort of average or max pooling to downsample the data, but then we decided to go for the 1-D convolutional layer instead, as the 1-D convolution is learnable and therefore we wouldn't be making any assumptions about which parts of the data were useful, letting the network figure that out for itself. We also experimented with using an attention layer, but ultimately found that our best model did not use the attention layer (however, the results were comparable with and without the attention). We used an augmented form of our base model: 
\begin{itemize}
    
\item 1-D Convolution (15 filters, stride 4, height 4), LSTM hidden size of 32, followed by two dense layers with 64 outputs, one dense layer with one output, using dropout with 0.2 probability and LeakyReLU between the dense layers, ending with no activation function. 
\item Table 4 shows the training, validation, and test MAPE error for this model, with Figure 7 visualizing these results.


\end{itemize}

\section{Results}

\subsection{Classification}

The entire 124 battery dataset was used to evaluate the model. The data was partitioned as followed: 79 train, 25 validation, 20 test. The dataset was partitioned randomly. Each model was trained on 79 batteries, and the first 5 cycles of the battery were inputted as a sequence into the model. At the end of our model search, we tested the best performing model on the 20 test batteries. Figure 5 shows the training and validation accuracy over 500 epochs for this model.


\begin{deluxetable}{l c c c}[H]
\tablecaption{Classification accuracy results.}
\tablecolumns{4}
\tablehead{
\colhead{} & 
\colhead{Train} & 
\colhead{Validation} &
\colhead{Test} 
}
\startdata
Accuracy (\%) & 90.5 & 90.4 & 95.0  \\
\enddata
\end{deluxetable}

\begin{figure}[H]
\centering
\includegraphics[width=.9215\linewidth]{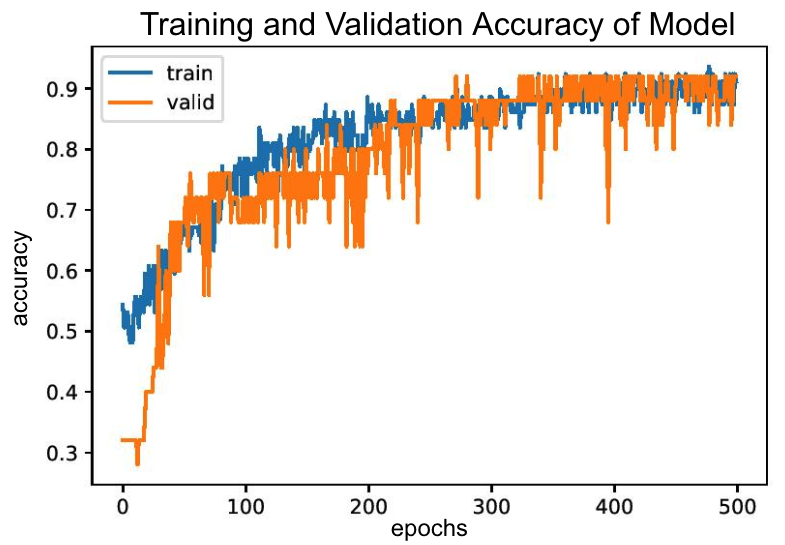}
\caption{Training and validation accuracy over 500 epochs with our simple, yet best performing model. The noise is likely due to our entire dataset only being 124 batteries: 79 train, 25 valid, 20 test.}
\end{figure}

After developing this simplified model, we hoped it would translate well to the test set, and results could be competitive with that of \cite{Severson}. Performance on the test set is visualized in Figure 6. 

\begin{figure}[H]
\centering
\includegraphics[width=\linewidth]{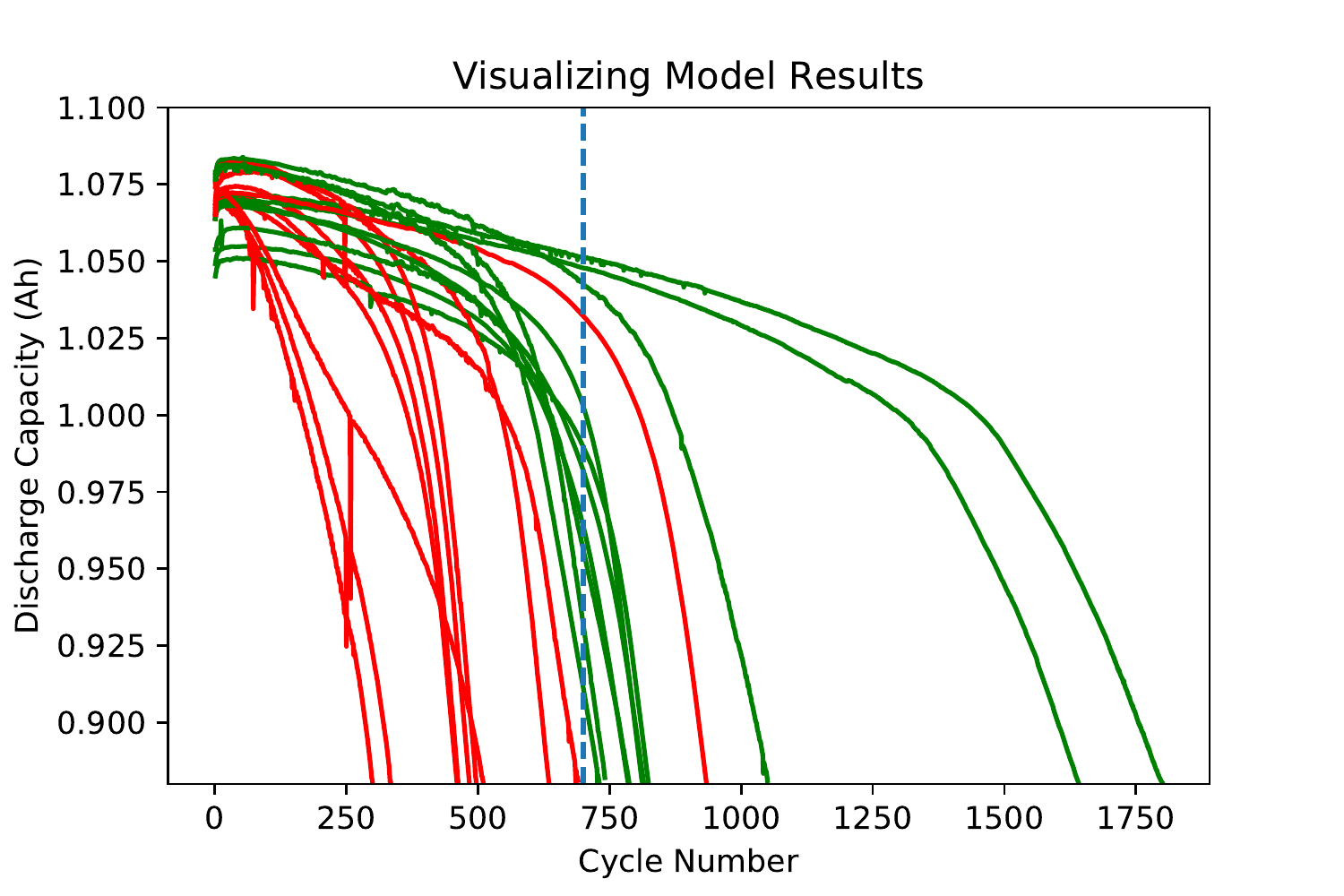}
\caption{Discharge capacity (\textit{Qd}) as a function of cycle number on our test set (95\% accuracy). When the \textit{Qd} lowers to the bottom of the figure, the battery is declared dead, and the cycle life is determined. The blue vertical line is at cycle 700. Each curve is a battery; looking at just the first 5 cycles, the green curves were predicted to last longer than 700 cycles, the red shorter. We see that 19 out of 20 batteries were correctly classified; one battery was classified as ``bad", but lasted longer than the cutoff.}
\end{figure}


\begin{deluxetable}{l c c c}[H]
\tablecaption{Test Error Comparison to Current Leading Model for Battery Cycle-life Classification.}
\tablecolumns{4}
\tablehead{
\colhead{} & 
\colhead{Our Model} & 
\colhead{Severson et al.} &
\colhead{Severson et al.} \\
\colhead{} & 
\colhead{} & 
\colhead{Primary} &
\colhead{Secondary}
}
\startdata
Accuracy (\%) & 95.0 & 92.7 & 97.5 \\
\enddata
\tablecomments{Our final model, which leveraged an LSTM and Attention to process the sequential data, compares to \cite{Severson}’s state of the art model on a held out test set of 20 batteries. While we acknowledge the limited size of the dataset, the model is certainly appears competitive with the current leading model for battery cycle degradation.}
\end{deluxetable}

\subsection{Prediction}

The entire 124 battery dataset was used to evaluate the model. The data was partitioned as followed: 79 train, 25 validation, 20 test. The dataset was partitioned randomly. Each model was trained on 79 batteries, and the first 100 cycles per battery were inputted as a sequence into the model. At the end of our model search, we tested the best performing model on the 20 test batteries. 

\begin{deluxetable}{l c c c}[H]
\tablecaption{Prediction accuracy results.}
\tablecolumns{4}
\tablehead{
\colhead{} & 
\colhead{Train} & 
\colhead{Validation} &
\colhead{Test} 
}
\startdata
MAPE (\%) & 10.0 & 16.0 & 12.5  \\
\enddata
\end{deluxetable}

Our final model achieved a 12.5\% test error, compared with Stanford's 9.1\%, using MAPE error for the comparison.
\begin{figure}[H]
\centering
\includegraphics[width=\linewidth]{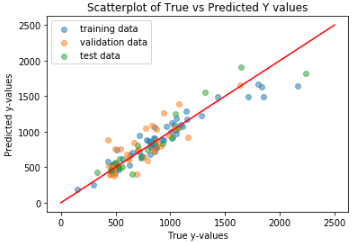}
\caption{Our final model for predicting the exact cycle-life of a battery finished with a training MAPE of 10\%, a validation MAPE of 16\%, and a test MAPE of 12.5\%.
}
\end{figure}


\begin{deluxetable}{l c c}[H]
\tablecaption{Test Error Comparison to Current Leading Model for Battery Cycle-life Prediction.}
\tablecolumns{3}
\tablehead{
\colhead{} & 
\colhead{Our Model} & 
\colhead{Severson et al.}
}
\startdata
Test MAPE (\%) & 12.5 & 9.1 \\
\enddata
\tablecomments{Our final model, which leveraged a 1-D convolutional layer plus an LSTM to process the sequential data compares to Stanford’s state of the art model on a held out test set of 20 batteries. While we acknowledge the limited size of the dataset, our model is comparable to the current leading model for battery cycle-life prediction.}
\end{deluxetable}

\section{Conclusion}
By the end of the evolutionary process, we had tried various model architectures with different combinations of hyperparameters, all in an effort to try to tackle the problem of predicting how a lithium-ion battery will last before it reaches approximately 80\% of its original capacity, rendering it ``dead." Simply passing in the most amount of data possible and hoping the model would learn was not a viable strategy; performance was greatly improved by filtering and reducing the data. Further, we learned that the most complex model doesn't necessarily perform the best---we tried many different, more complex iterations of our baseline model, and ultimately found that the simpler models actually turned out to perform best.  Additionally, we observed the effectiveness of the concept of Attention being used outside the scope of NLP. It applied to our task, although we found that it did not help all that much compared to just using pure LSTMs as it does in NLP tasks. We postulate this to be due to the fact that the relationship between each of the cycles of the battery are more temporally linear than words in a sentence, which can refer to things far in the past, as well as be related to multiple other words.

Our biggest goal of this paper was to create a model competitive with the current state of the art accuracies, approaching the problem from a purely data-driven standpoint, hoping to use deep learning to learn the patterns in the sequences of data that the Stanford team engineered by hand. For the classification task, were able to achieve very competitive results, with validation accuracies above 90\%, and a test accuracy of 95\%, compared to the 97.5\% test accuracy of the current leading model \citep{Severson}. For the prediction task, we were also able to achieve competitive results, with a test MAPE error of 12.5\% as compared with a 9.1\% error achieved by the current leading model \citep{Severson}. We believe a larger dataset would allow for more refined  experimentation, allowing us to improve upon current results.





\end{document}